%% file: main.tex
\documentclass{article}
\usepackage{preprint,times}
\usepackage[letterpaper,textwidth=6.3in,top=1.02in,bottom=1.04in,
            includefoot,footskip=14pt]{geometry}
\usepackage{amsmath,amssymb,booktabs,graphicx,microtype}
\input{math_commands.tex}

\usepackage{hyperref}
\makeatletter
\let\TaReD@natlinkstart\hyper@natlinkstart
\let\TaReD@natlinkend\hyper@natlinkend
\renewcommand*{\hyper@natlinkstart}[1]{\leavevmode\hbox\bgroup\TaReD@natlinkstart{#1}}
\renewcommand*{\hyper@natlinkend}{\TaReD@natlinkend\egroup}
\renewcommand*{\hyper@natlinkbreak}[2]{\hyper@natlinkend#1\hyper@natlinkstart{#2}}
\makeatother
\usepackage{url}
\usepackage{enumitem}
\usepackage{float}
\usepackage{capt-of}
\usepackage{algorithm}
\usepackage{algpseudocode}
\usepackage{needspace}
\renewcommand{\normalsize}{\fontsize{10}{11.95}\selectfont}
\let\small\normalsize
\let\footnotesize\normalsize
\title{TaReD: Tool-Aware Recursive Decomposition for Long-Horizon Tasks}
\author{%
Wei-Xiang Mao\textsuperscript{1,2}\thanks{Equal contribution.}\quad
Zhi-Kai Chen\textsuperscript{2,3}\footnotemark[1]\quad
De-Chuan Zhan\textsuperscript{2,3}\quad
Han-Jia Ye\textsuperscript{2,3}\thanks{Corresponding author.}\\[4pt]
{\normalfont\textsuperscript{1}Nanjing University, China}\\
{\normalfont\textsuperscript{2}National Key Laboratory for Novel Software Technology, Nanjing University, China}\\
{\normalfont\textsuperscript{3}School of Artificial Intelligence, Nanjing University, China}%
}
\preprintfinalcopy
\hypersetup{
    hidelinks,
    pdftitle={TaReD: Tool-Aware Recursive Decomposition for Long-Horizon Tasks},
    pdfauthor={Wei-Xiang Mao, Zhi-Kai Chen, De-Chuan Zhan, Han-Jia Ye}
}

\begin{document}
\raggedbottom
\maketitle
\fancyhead{}
\renewcommand{\headrulewidth}{0pt}
\begin{abstract}
\setlength{\parskip}{0pt}
Agents combine reasoning with tools to interact with external systems and complete real-world tasks. Early agents typically interleave reasoning and actions along a single execution chain. On complex tasks, this chain becomes unreliable because growing histories obscure intermediate dependencies and allow early planning errors to propagate. Recursively decomposing a complex task into smaller subtasks offers a natural solution, yet effective decomposition must account for the system's capabilities so that each subtask can be executed by the available tools. In realistic systems, however, tool libraries can be too large to expose in full. Injecting every tool description consumes substantial context while making relevant tools harder to retrieve and useful task boundaries harder to identify. We propose tool-aware recursive decomposition, which organizes tools by functional relationships into a hierarchy of capabilities. During execution, the agent discovers tools on demand and uses the hierarchy to recursively decompose a complex task into a subtask tree whose levels are aligned with the capabilities required at each stage. Experiments on complex real-world tasks show that the proposed method improves end-to-end task success rate by up to 40 percentage points over the compared baselines. The implementation of TaReD is available on GitHub: \url{https://github.com/WeiXiang-Mao/TaReD}.
\end{abstract}
\vspace{-0.5em}

\setlength{\parskip}{2.7pt}
\input{sections/1_introduction}
\setlength{\parskip}{2.7pt}
\input{sections/2_related_work}
\input{sections/3_problem_and_metrics}
\input{sections/4_method}
\begingroup
\setlength{\parskip}{2.7pt}
\input{sections/5_experimental_setup}
\input{sections/6_discussion_and_conclusion}
\endgroup

\clearpage
\bibliography{references}
\bibliographystyle{preprint}

\clearpage
\appendix
\input{sections/A_appendix}
\clearpage
\input{sections/B_execution_trace}
\end{document}

%% file: math_commands.tex
\newcommand{\T}{\mathcal{T}}
\newcommand{\G}{\mathcal{G}}
\newcommand{\Htree}{\mathcal{H}}
\newcommand{\traj}{\tau}

%% file: sections/1_introduction.tex
\section{Introduction}

Agents combine language-based reasoning with executable tools to retrieve information~\citep{karpas2022mrkl,nakano2021webgpt} and change state in external environments~\citep{yao2023react,zhou2024webarena}. Tool interfaces extend an agent's capabilities beyond its language context~\citep{wang2025toolgen}, but also introduce additional planning requirements: the agent must identify which tool is relevant~\citep{schick2023toolformer,shen2023hugginggpt}, construct valid arguments~\citep{patil2023gorilla,tang2023toolalpaca,chen2026schemafill}, order calls whose results and state changes may influence one another~\citep{qin2023toolllm,chen2026peakbench}, and maintain \mbox{those dependencies across a long sequence of operations.}

Early agents typically interleave reasoning and actions along a single execution chain~\citep{yao2023react,paranjape2023art}. In this chain-based style, the agent repeatedly reasons about the current state, selects and invokes a tool, observes its result, and then continues from the updated state~\citep{huang2023innermonologue}. As tasks become more complex, the chain must carry more intermediate states and dependencies, making it harder to maintain and recover; an early planning or execution error can propagate through later steps and \mbox{reduce end-to-end success~\citep{liu2024agentbench}.} 

Task decomposition addresses the limitations by breaking a complex task into an ordered sequence of subtasks~\citep{zhou2023leasttomost} with explicit dependencies~\citep{khot2022decomposed}. Early decomposition methods generally derive these subtask boundaries from task semantics, such as goals, entities, and procedural structure~\citep{wang2023planandsolve}. However, semantic decomposition alone does not ensure that each subtask corresponds to an operation the current system can execute: it may introduce a subtask with no directly \mbox{executable capability~\citep{huang2022zeroshot} or omit a necessary stage.}

\begin{figure}[t]
    \centering
    \makebox[\linewidth][c]{\includegraphics[width=5.39in]{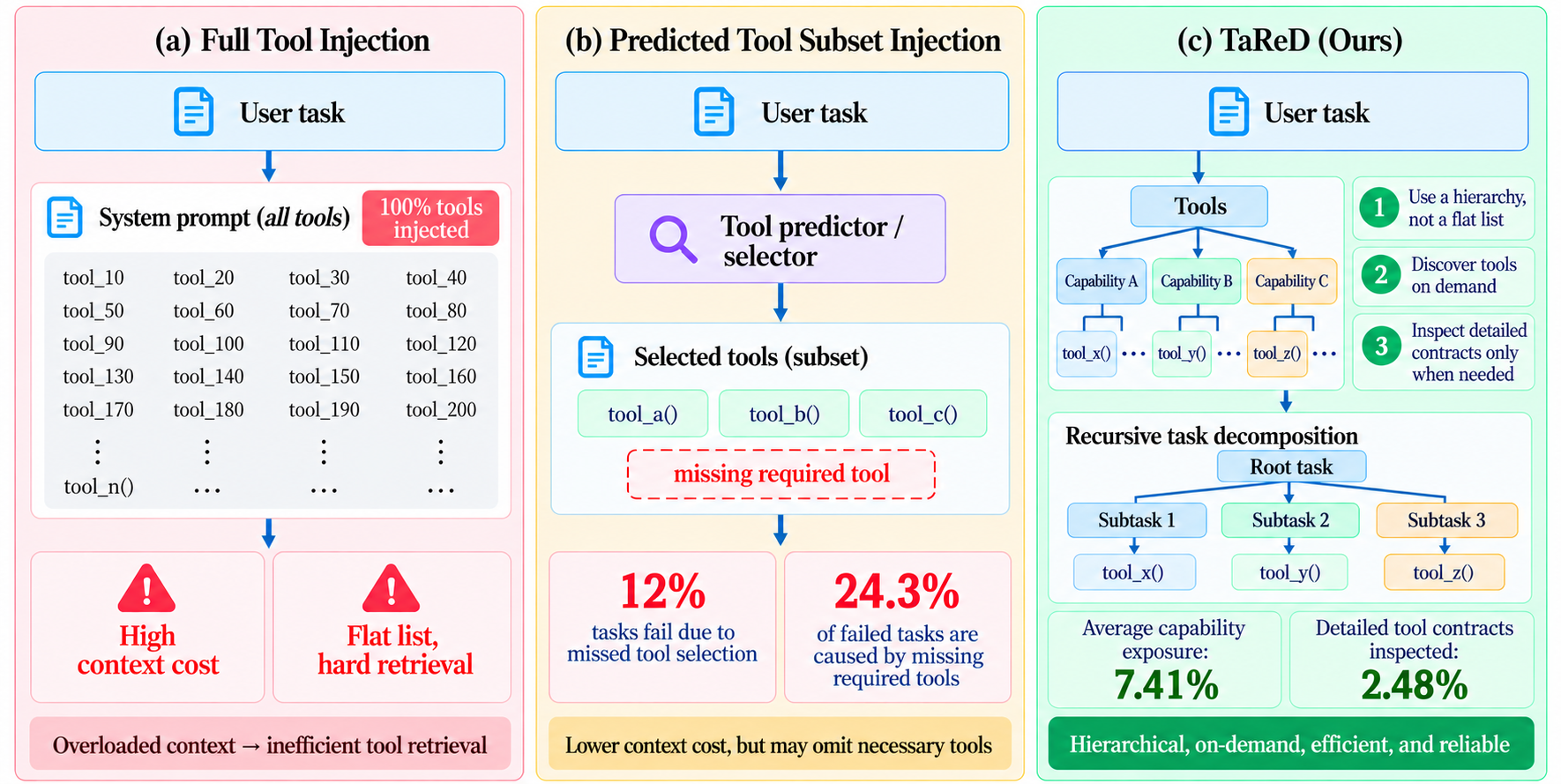}}
    \vspace{-5pt}
    \caption{Motivation for tool-aware recursive decomposition. (a) Injecting descriptions and contracts of the full library consumes substantial context and distracts selection of task-relevant tools. (b) A fixed predicted subset can omit necessary tools and offers little structural guidance for dividing a task into capability-aligned subtasks. (c) TaReD navigates a capability hierarchy to retrieve relevant information on demand, guide recursive decomposition toward executable stages, and inspect detailed contracts only when needed.}
    \label{fig:motivation}
\end{figure}

A high-quality task decomposition should align with the execution environment's capabilities~\citep{ahn2023saycan,singh2023progprompt}, so that every subtask can be carried out by corresponding operations available in that environment~\citep{song2023restgpt}. Achieving this requires information about system capabilities~\citep{karpas2022mrkl}. The challenge is to expose enough capability information~\citep{li2023apibank} to make subtasks executable without overwhelming the planner. As tool libraries continue to grow in realistic tasks, this challenge becomes increasingly pronounced: injecting the descriptions and contracts of all tools consumes substantial context~\citep{hao2023toolkengpt}, while the volume of tool information can distract the planner from selecting tools relevant to the task. Predicting a small, task-relevant tool subset~\citep{qin2023toolllm} and exposing only that subset to the planner partially alleviates this burden. However, this approach may omit necessary tools~\citep{shi2025toolret} and still presents tool information as a flat list, providing little structural guidance for task decomposition.

We propose \emph{TaReD} (Tool-Aware Recursive Decomposition), which organizes tools by functional relationships into a hierarchy of capabilities. When performing a task, TaReD navigates the hierarchy on demand, uses the discovered capability region to recursively split a task into functional stages, and binds an exact tool contract only when a stage is ready to execute. TaReD only retrieves the tool information needed for each task, avoiding both full-library context injection and the omission of necessary tools caused by a fixed subset. The hierarchy's functional relations also provide structural guidance for dividing the task into capability-aligned stages (Figure~\ref{fig:motivation}).

To assess the effectiveness of our method, we evaluate TaReD on complex interactive tasks in AppWorld~\citep{trivedi2024appworld}, comparing it with representative recursive~\citep{prasad-etal-2024-adapt}, linear~\citep{yao2023react}, and plan-then-execute~\citep{wang2023planandsolve} strategies. All implementations use matched models, budgets, verifiers, and service conditions. Experimental results show that TaReD improves end-to-end task success rate by up to 40 percentage points over the compared baselines. Our contributions are summarized as follows:

\begin{itemize}[leftmargin=1.2em,itemsep=0.2ex,topsep=0.2ex]
    \item We propose TaReD (Tool-Aware Recursive Decomposition), which organizes a large tool library into a hierarchy and tightly couples tool discovery with task decomposition.
    \item We implement an architecture and execution protocol for tool-aware recursive decomposition that fully realizes the proposed algorithm and achieves strong performance in task execution.
    \item We evaluate TaReD on complex, long-horizon tasks and show a clear improvement in end-to-end \mbox{performance over the compared baselines.}
\end{itemize}

%% file: sections/2_related_work.tex
\section{Related Work}

\paragraph{Task decomposition and hierarchical planning.} Chain-of-thought agents~\citep{wei2022chain} execute a task as a single sequence of interleaved reasoning and actions~\citep{yao2023react}. But as task complexity grows, long action chains strain state tracking and allow early errors to propagate. Task-decomposition methods address this limitation~\citep{prasad-etal-2024-adapt} by breaking complex tasks into interdependent subproblems~\citep{zhou2023leasttomost,xu2023rewoo} and solving them in an order~\citep{wang2023planandsolve} that respects their dependencies~\citep{khot2022decomposed,kim2024llmcompiler}. Early semantic-driven studies decomposed tasks based on their goals, entities~\citep{press2022selfask}, or procedural structure. However, semantic decomposition alone does not account for which operations the environment can actually execute. A subtask may therefore fail to specify an executable sequence, motivating the use of tool information to ground decomposition.

\paragraph{Tool use and large tool libraries.} Early tool-using research~\citep{karpas2022mrkl} studied how agents decide when to invoke tools~\citep{schick2023toolformer} and how to incorporate their results into ongoing reasoning~\citep{yao2023react}. In these settings, the available tools and their descriptions were typically provided in the agent's context~\citep{lu2023chameleon}. However, as the scale of tools increases, it is impractical to include the full tool information in the context~\citep{hao2023toolkengpt}. To avoid overwhelming the planner with lengthy tool information, some methods first predict or retrieve a task-relevant subset and plan over that subset~\citep{qin2023toolllm}. Although selection based on prediction or retrieval reduces irrelevant choices, it can omit a necessary tool even when the full library contains a solution~\citep{shi2025toolret,zheng2024toolrerank}. Moreover, the selected tools are typically presented as a flat list, which provides little guidance about capability relations or how to organize subtasks around capability boundaries.

\paragraph{Contributions of this work.} Our work organizes tools by functional relationships into a hierarchy of capabilities and queries this hierarchy on demand during execution. This avoids placing the full tool library in the context while retaining on-demand access to every tool in the library, reducing the risk of omitting a necessary tool. The hierarchy also exposes relations among capabilities and provides structural guidance for decomposing tasks into subtasks aligned \mbox{with the operations available in the environment.}

%% file: sections/3_problem_and_metrics.tex
\section{Preliminary}

We study how an agent can leverage the extensive tool information available in its execution environment to decompose complex tasks effectively and improve long-horizon task execution. A successful execution should achieve the task objective while driving the environment from its initial state to a desired final state.

\subsection{Task and environment}
We consider a long-horizon task specified by an instruction $x$ and an initial environment state $s_0$. The agent interacts with the environment through a library of executable tools. Each tool $T_i$ is described by a contract
\begin{equation}
T_i=(\mathsf{In}_i,\mathsf{Out}_i,\mathsf{Pre}_i,\mathsf{Eff}_i),
\end{equation}
\looseness=-1\textls[-10]{where $\mathsf{In}_i$ and $\mathsf{Out}_i$ specify its input and output schema, while $\mathsf{Pre}_i$ and $\mathsf{Eff}_i$ describe its preconditions and state-changing effects. Calling a tool with arguments $z_t$ produces an observation $o_t$ and updates the environment according to}
\begin{equation}
(o_t,s_{t+1})=\mathsf{Execute}(T_i,z_t,s_t).
\end{equation}
An execution trajectory is a sequence $\traj=((T_{i_t},z_t,o_t))_{t=1}^{L}$. The task is completed when a goal \mbox{predicate $\mathsf{Done}(x,s_L)$ is satisfied.}

\subsection{Tool library and capability hierarchy}
Let $\T=\{T_1,\ldots,T_n\}$ denote the complete tool library. The library may be too large to expose in the context at once, so we organize tools by functional category into a tree:
\begin{equation}
\G=(\mathcal{V},\mathcal{E}),
\end{equation}
where internal nodes summarize capability regions and leaf nodes correspond to tools with complete execution contracts. Each node $v\in\mathcal{V}$ is associated with a capability description $\mathsf{cap}(v)$ and, when $v$ is a leaf, a tool contract $T_i$. A path from the root to a leaf therefore represents progressively more \mbox{specific information about an executable capability.} 

\subsection{Recursive task decomposition}
We represent a decomposition as a rooted, ordered task tree whose nodes correspond to functional stages. Specifically, the tree and a stage $h$ are represented as
\begin{equation}
\Htree=(\mathcal{N},\mathcal{E}_h),\qquad h=(q_h,c_h,r_h),\quad h\in\mathcal{N}.
\end{equation}
Here, $q_h$ is the stage goal, $c_h$ contains the relevant task context, and $r_h\in\mathcal{V}$ denotes the capability region expected to carry out the stage. An internal node is recursively decomposed into an ordered sequence of child stages, $h\rightarrow(h_1,\ldots,h_k)$, while a leaf stage is ready to bind one or more concrete tool contracts and execute.

\subsection{Problem objective}
Let $\mathsf{Align}(\Htree,\T)$ denote the condition that every leaf in the recursive task tree has a corresponding tool in the execution environment and that each tool call satisfies its precondition at the time of execution. Given $(x,s_0,\G)$, we seek a task tree $\Htree$ such that $\mathsf{Align}(\Htree,\T)$ holds and the resulting trajectory satisfies $\mathsf{Done}(x,s_L)$. We therefore formulate the planning objective as
\begin{equation}
\max_{\Htree,\traj}\; \Pr\!\left[\mathsf{Done}(x,s_L)\right]
\quad\text{s.t.}\quad
\mathsf{Align}(\Htree,\T)
\end{equation}
The objective maximizes the probability that the final environment state satisfies the task goal, while the alignment constraint ensures that the recursive decomposition is consistent with the tool library available \mbox{in the execution environment.}

%% file: sections/4_method.tex
\section{TaReD: Tool-Aware Recursive Decomposition}
Existing research on agents still faces challenges when executing long-horizon tasks in environments with large tool libraries. Injecting the full tool library into the context can overwhelm the planner, whereas predicting or retrieving a smaller subset may omit necessary tools and offers little guidance for structuring subtasks. TaReD addresses these limitations by organizing tools into a capability hierarchy and discovering them on demand, keeping the context focused while preserving access to the full library and guiding decomposition around capability relations.

\begin{figure}[!htb]
    \centering
    \makebox[\linewidth][c]{\includegraphics[width=5.39in]{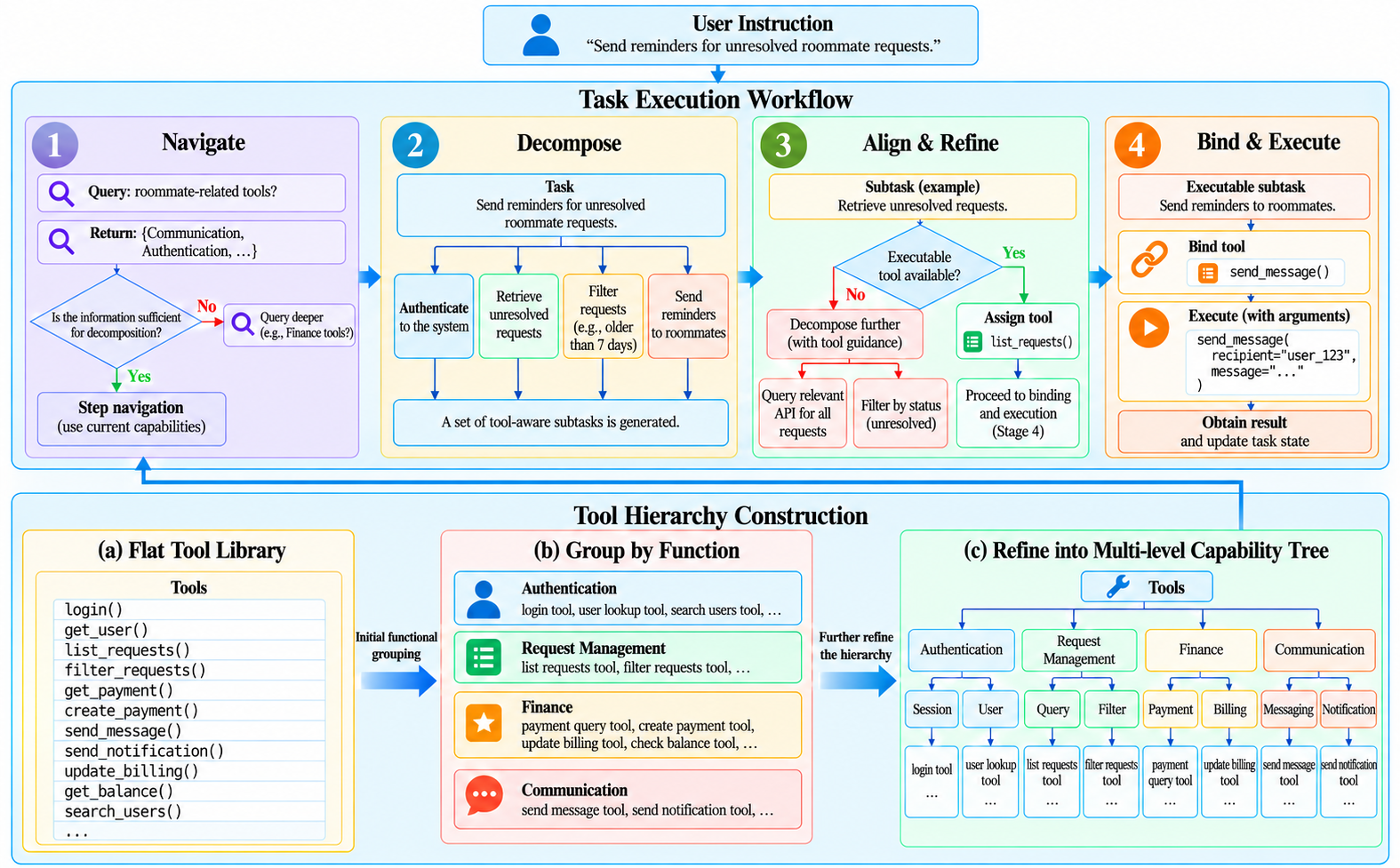}}
    \vspace{-5pt}
    \caption{Overview of TaReD. Bottom: a flat tool library is grouped by function and refined into a multilevel capability tree. Top: the agent navigates this hierarchy, decomposes and refines subtasks using capability information, then binds arguments and executes tools when a stage is ready to execute. Returned results \mbox{update the task state for subsequent stages.}}
    \label{fig:tared-overview}
\end{figure}

\subsection{Overall workflow}
TaReD organizes a large tool library into a tree according to functional relationships among tools. During execution, the agent starts at the general root and navigates downward as needed to discover increasingly specific capabilities and tool information; information about irrelevant branches is not added to the context. Starting from the original task, the agent performs this tool-awareness process until it has enough capability information to identify functional stages, then decomposes the task into an ordered set of subtasks and processes them depth first. For each subtask, the agent first decides whether it is directly executable. If it remains complex, TaReD repeats tool awareness and decomposition recursively. If it is simple enough to execute, the agent plans the required tool calls, retrieves their detailed contracts, binds the necessary arguments, and invokes the tools. The resulting values and environment state are stored for use by subsequent dependent subtasks. Execution terminates when no pending task nodes remain or when the agent determines that the available capabilities are insufficient to continue the task.

\subsection{Hierarchical capability representation}
TaReD organizes the complete tool library into a rooted hierarchy
\(\G=(\mathcal{V},\mathcal{E})\). The hierarchy contains a root node for the complete library, internal capability nodes at different levels of granularity, and leaf nodes corresponding to executable tools. Internal nodes represent functionally related capability regions and provide navigation rather than execution. Each internal node is associated with a compact capability card
\begin{equation}
v=(\mathsf{name},\mathsf{summary},\mathsf{children},\mathsf{scope}),
\end{equation}
which summarizes the operations supported by that region, the kind of information it consumes and produces, and the boundary of the capability it covers. Child links expose progressively more specific functional regions. A leaf represents a concrete executable operation and provides the exact contract needed \mbox{when that operation is ready to run.}

The hierarchy is therefore more than a flat index of tool names. Its internal relations expose functional boundaries that can be used to scope a subtask: a high-level node identifies what kind of operation is needed, while progressively deeper nodes provide the specificity required to choose an executable tool. By making these relations explicit, the hierarchy represents the execution environment's capability structure and provides guidance for task decomposition.

\subsection{On-demand capability navigation}
For a task node $h=(q_h,c_h,r_h)$, where $q_h$ is the current goal, $c_h$ is the available context, and $r_h$ is its expected capability region, the controller navigates the capability hierarchy using the goal and the state of the task:
\begin{equation}
C=\mathsf{Navigate}(q_h,c_h,r_h;\G).
\end{equation}
Starting from the expected region $r_h$, the controller queries the capability hierarchy level by level, gathering information relevant to the current goal $q_h$ and context $c_h$. It continues until the retrieved information is sufficient to plan the current task node. If the expected region does not provide all the capability information needed for planning, the controller searches other regions of the hierarchy and uses their information to fill the gaps. Only information from visited nodes is added to the context.

On-demand navigation keeps irrelevant tool information out of the context and focuses the agent on capabilities needed for the current task. Because the controller can search the complete hierarchy, necessary tools are not omitted by an initially predicted subset.

\subsection{Capability-aligned recursive decomposition}
Once the available capability information is sufficient to reason about the current goal, TaReD decides whether that goal can be handled as a compact execution stage. If so, it proceeds directly to the corresponding tool calls, without adding another task-decomposition level. When further decomposition is needed, TaReD uses the acquired capability information to represent the current task as an ordered sequence of child tasks. Let $h=(q_h,c_h,r_h)$ denote the current task node. Its recursive refinement is
\begin{equation}
\mathsf{Refine}(h)=
\begin{cases}
\mathsf{Exec}(h), & \mathsf{Direct}(h)=\mathrm{True},\\
\left(\mathsf{Refine}(h_j)\right)_{j=1}^{k}, & \mathsf{Direct}(h)=\mathrm{False},
\end{cases}
\qquad h_j=(q_j,c_j,r_j),\quad k>1,
\end{equation}
where $\mathsf{Direct}(h)\in\{\mathrm{True},\mathrm{False}\}$ indicates whether the current goal and context, together with the observed capability information, support a compact executable stage. When $\mathsf{Direct}(h)=\mathrm{False}$, TaReD forms $k>1$ child tasks $h_1,\ldots,h_k$ whose goals $q_1,\ldots,q_k$ capture the capability-grounded functional parts of $h$. Let $\mathcal{D}_h$ denote the dependency relation among these children, where $(h_i,h_j)\in\mathcal{D}_h$ means that $h_j$ needs an output produced by $h_i$. For each dependency $(h_i,h_j)\in\mathcal{D}_h$, let $B_{ij}$ denote the input bindings that pass only the values required from $h_i$ to $h_j$:
\begin{equation}
\mathcal{D}_h\subseteq\{(h_i,h_j)\mid 1\leq i<j\leq k\},
\qquad
c_j=c_h\cup\{B_{ij}\mid (h_i,h_j)\in\mathcal{D}_h\}.
\end{equation}
TaReD processes the ordered children depth first: it fully refines and handles $h_1$ before moving to $h_2$, and so on. A child that is itself decomposed is handled by the same rule before its parent advances to the next sibling. Each $B_{ij}$ references only the output value required from $h_i$, rather than merging that predecessor's full context or output record into $c_j$. The scheduler dispatches a child only after its declared predecessors are closed and its bindings can be resolved to established values. After a child returns, the parent receives its status, summary, and outputs. If the child completed and the committed layer permits automatic advancement, the scheduler dispatches the next pending sibling in order; otherwise, or when a child fails, is uncertain, or requires review, control returns to the parent for review, repair, or replanning. The parent judges whether its goal is satisfied from the actual results and may declare completion when no child obligations remain open and required outputs are available.

The capability hierarchy makes the execution environment's functional structure explicit and guides task decomposition toward capabilities the environment can support. Because navigation proceeds from broad functions to specific operations, TaReD can divide a task according to the capability represented by an intermediate node, without requiring every decomposition decision to select individual tools. This allows the planner to control subtask granularity at the level of functional stages and refine toward concrete operations only when needed. On-demand navigation also keeps each planning context focused on tool information relevant to the task, helping the agent devote its attention to decomposition rather than unrelated capabilities.

\subsection{Delayed contract binding and execution}
TaReD delays exact contract binding until the current task is judged directly executable. The controller then plans an ordered sequence of one or more tool calls, $P_h=(a_1,\ldots,a_m)$. Before each call, it retrieves the corresponding contract $d_i$ and binds arguments from that contract, the stage context, and the values already stored in $\mathcal{M}$. Calls execute in the planned order, with each result and updated environment state available to subsequent calls. For a stage entered with state $s_0$ and store $\mathcal{M}_0$, where these denote the current state and value store at the start of this stage rather than the task's initial state or an empty store, the execution and result recording at step $i$ are
\begin{align}
z_i&=\mathsf{Bind}(d_i,c_h,\mathcal{M}_{i-1}),\quad
(o_i,s_i)=\mathsf{Execute}(a_i,z_i,s_{i-1}), \label{eq:stage-bind-execute}\\
\mathcal{M}_i&=\mathcal{M}_{i-1}\cup\{(k_i,o_i,\mathsf{prov}(o_i))\},
\qquad i=1,\ldots,m. \label{eq:stage-store-result}
\end{align}
Here, $z_i$ is a contract-valid argument assignment, $o_i$ is the returned observation, $s_i$ is the resulting environment state, and $k_i$ is a stable reference to the stored value and its available provenance. Later calls in the same stage and dependent subtasks can use these references to retrieve required values.

This design keeps detailed parameter schemas out of earlier decomposition decisions while grounding every executed call in its exact contract. It also lets one task stage coordinate multiple sequential operations without forcing each operation into a separate recursive subtask. The referenced value store supports reliable handoffs between calls and subtasks without copying full earlier responses into the context; provenance remains available for tracing, while values can still be used when provenance is unavailable. Together, delayed binding and reference-based handoffs preserve state and dependencies while keeping execution focused on the information each call needs.

\subsection{Algorithmic summary}
Algorithm~\ref{alg:tared} expresses TaReD as a recursive procedure over task nodes. At each call, $h$ denotes the current task, and $Q$ is the ordered set of subtasks produced by decomposing it; $\mathcal{D}$ records their dependencies, and $B$ contains their input bindings. Sufficiency and direct-execution checks are planner judgments grounded in observed capability information. Each recursive call handles one task, and its subtask results return \mbox{to that current task for review and further planning.}

\begingroup
\floatstyle{plain}
\restylefloat{algorithm}
\begin{algorithm}[!t]
\hrule\vspace{2pt}
\refstepcounter{algorithm}
\noindent\textbf{Algorithm \thealgorithm} TaReD: Tool-Aware Recursive Decomposition\par\nobreak
\vspace{2pt}\hrule\nobreak
\label{alg:tared}
\begin{algorithmic}[1]
\Require Original task $x$, initial context $c_0$, environment state $s_0$, capability hierarchy $\G$
\Ensure Root status $y_0$, final environment state $s$, referenced value store $\mathcal{M}$
\State Initialize $h_0=(x,c_0,\operatorname{root}(\G))$, $\mathcal{M}\gets\emptyset$, and $s\gets s_0$
\State $(y_0,s,\mathcal{M})\gets\Call{HandleNode}{h_0,s,\mathcal{M},\G}$
\State \Return $(y_0,s,\mathcal{M})$
\Statex
\Function{HandleNode}{$h,s,\mathcal{M},\G$}
    \nopagebreak[4]
    \State $C\gets\Call{Navigate}{q_h,c_h,r_h,\G}$
    \If{$\Call{Sufficient}{h,s,\mathcal{M},C}=\mathrm{False}$}
        \State \Return $(\mathsf{insufficient},s,\mathcal{M})$
    \EndIf
    \If{$\Call{Direct}{q_h,c_h,s,\mathcal{M},C}=\mathrm{True}$}
        \State $P_h\gets\Call{PlanCalls}{q_h,c_h,s,C}$
        \ForAll{$a\in P_h$ in the planned order}
            \State $d\gets\Call{InspectContract}{a}$; bind arguments $z$ from $d$, $c_h$, and $\mathcal{M}$
            \State $(o,s)\gets\Call{Execute}{a,z,s}$
            \State Assign fresh reference $k$ to $o$ and set $\mathcal{M}[k]\gets(o,\mathsf{prov}(o))$
            \State Expose $k$ as an output reference for dependent subtasks
        \EndFor
        \State $y_h\gets\Call{AssessStage}{h,s,\mathcal{M}}$
        \State \Return $(y_h,s,\mathcal{M})$
    \EndIf
    \State $(Q,\mathcal{D},B)\gets\Call{Decompose}{q_h,c_h,s,C}$
    \State $\mathsf{review\_required}\gets\mathrm{False}$
    \While{pending subtasks remain or the current task requires review}
        \State $h_j\gets$ first pending depth-first subtask with completed $\mathcal{D}$-predecessors and resolvable $B$
        \If{$h_j=\varnothing$ or $\mathsf{review\_required}=\mathrm{True}$}
            \State $(Q,\mathcal{D},B,\rho)\gets\Call{ReviewOrRevise}{h,Q,\mathcal{D},B,s,C}$
            \If{$\rho=\mathsf{stop}$}
                \State \Return $(\mathsf{insufficient},s,\mathcal{M})$
            \EndIf
            \State $\mathsf{review\_required}\gets\mathrm{False}$
        \Else
            \State Resolve $h_j$'s inputs using $B$, retrieving referenced predecessor values from $\mathcal{M}$
            \State $(y_j,s,\mathcal{M})\gets\Call{HandleNode}{h_j,s,\mathcal{M},\G}$
            \State Record $y_j$ and its output references in $Q$
            \If{$y_j\neq\mathsf{completed}$}
                \State $\mathsf{review\_required}\gets\mathrm{True}$
            \EndIf
        \EndIf
    \EndWhile
    \State $y_h\gets\Call{AssessTask}{h,s,\mathcal{M}}$
    \State \Return $(y_h,s,\mathcal{M})$
\EndFunction
\end{algorithmic}
\par\vspace{2pt}\hrule
\end{algorithm}
\endgroup

Appendix~\ref{sec:illustrative-trace} walks through an illustrative task to show how capability discovery, recursive subtask dependencies, and delayed contract binding interact during execution.

%% file: sections/5_experimental_setup.tex
\section{Experiments}
\subsection{Experimental Setup}
\noindent\textbf{Environment and evaluation.} AppWorld is an interactive environment with long-horizon tasks and a large tool library of 457 tools spanning nine applications~\citep{trivedi2024appworld}. Its tasks are organized into scenarios, each containing several variants with a shared underlying objective and consistent task type. It provides an official programmatic verifier for checking the resulting environment state. We evaluate the most challenging tasks in its \emph{Challenge set (Test-C)}, using GPT-5.6 Luna for all methods and a 600-second limit per episode. Other shared run settings include a fixed random seed of 42, a 100-step episode cap and temperature-0 generation.

\noindent\textbf{Compared methods.} We compare four methods: \emph{TaReD}, \emph{ADAPT}, \emph{ReAct}, and \emph{Plan-and-Solve}. We use the official AppWorld implementations of ReAct and Plan-and-Solve, while retaining ADAPT's official architecture and adapting it to the AppWorld Challenge set. TaReD combines hierarchical tool awareness with recursive task decomposition and delays exact tool-contract binding until execution. ADAPT follows an executor-first recursive strategy, decomposing a task when direct execution cannot complete it~\citep{prasad-etal-2024-adapt}. ReAct interleaves reasoning and tool calls in a single linear trajectory without recursive task decomposition~\citep{yao2023react}. Plan-and-Solve follows the two-stage design of first generating a complete natural-language plan and then executing the plan sequentially~\citep{wang2023planandsolve}. All three baselines use a one-shot Predictor \mbox{to select a task-specific subset of tools.}

\noindent\textbf{Metrics.} We report AppWorld's Task Goal Completion (TGC), Scenario Goal Completion (SGC), and an Average Task Verification Score (ATVS). TGC is the fraction of individual tasks for which all official verifier checks pass. SGC is the fraction of scenarios for which all task variants in the scenario pass, measuring consistency across related task formulations. Let $p_i$ and $m_i$ denote the number of verifier checks passed and performed for task $i$, respectively. The ATVS is
\begin{equation}
S=\frac{100}{N}\sum_{i=1}^{N}
\begin{cases}
p_i/m_i, & \text{if the verifier ran, }m_i>0,\text{ and }p_i>0,\\
0, & \text{otherwise},
\end{cases}
\end{equation}
where $N$ is the number of evaluated tasks. This normalization sets the maximum ATVS to 100.

\subsection{Main Results and Analysis}
\begingroup
\setlength{\intextsep}{6pt plus 2pt minus 2pt}
\begin{table}[!htbp]
\centering
\caption{Results on the AppWorld Challenge set (Test-C). Task Goal Completion (TGC) requires all official verifier checks to pass for an individual task, Scenario Goal Completion (SGC) requires all variants in a scenario to pass, and Average Task Verification Score (ATVS) reports normalized verifier progress with partial credit. TaReD achieves the highest value on all three metrics in the adopted run set.}
\label{tab:main-results}
\vspace{4pt}
\begin{tabular}{lccc}
\toprule
Method & TGC & SGC & ATVS \\
\midrule
ReAct~\citep{yao2023react} & 51.3\% & 26\% & 66.81 \\
Plan-and-Solve~\citep{wang2023planandsolve} & 29.3\% & 16\% & 60.60 \\
ADAPT~\citep{prasad-etal-2024-adapt} & 46.7\% & 22\% & 66.01 \\
\textbf{TaReD (ours)} & \textbf{69.3\%} & \textbf{48\%} & \textbf{80.28} \\
\bottomrule
\end{tabular}
\end{table}
\endgroup

Overall, TaReD performs best among the four methods in Table~\ref{tab:main-results} across individual-task completion, scenario-level consistency, and partial-credit verification. TaReD's advantage over ADAPT follows from the interaction between recursive control and tool awareness. ADAPT recursively plans inside a one-shot, flat tool subset, so omitting a necessary inspection, lookup, or verification tool can leave the task impossible to complete. Moreover, the flat tool list provides little guidance about capability relations or boundaries, making it harder to organize the task into well-grounded functional stages. TaReD instead uses the capability hierarchy to organize the task into functional stages, expands discovery when the current region is insufficient, and postpones exact contract binding until a stage is ready to execute. The combination of effective tool awareness and a well-designed recursive framework enables higher-quality task decomposition and reduces issues that may arise during recursive execution.

ReAct follows a single linear reasoning process with the same flat tool-selection setting. It can avoid child scheduling, dependency handoff, and recursive-plan validation errors that affect ADAPT. However, its linear execution continually accumulates interaction history while providing little structure for organizing the task into distinct stages. Like ADAPT, ReAct also faces the risk of missing necessary tools. TaReD combines the staging benefits of recursion with capability-guided, on-demand discovery, which is consistent with its better performance in this evaluation. Plan-and-Solve provides another non-recursive planning comparison. It has the lowest TGC, SGC, and ATVS. In addition to potentially missing necessary tools under the same flat Predictor-selected view, Plan-and-Solve commits to a complete plan before execution, leaving limited scope to revise it in response to execution-time problems; consequently, an early error can disrupt subsequent steps.

\subsection{Ablation Studies}
\noindent\textbf{Tool Awareness within the Recursive Framework.}
We compare four tool-awareness configurations within TaReD's recursive framework. \emph{Predictor subset} selects a fixed flat tool subset once before execution. \emph{Full tool injection} places all 457 short tool descriptions in the context and retrieves detailed contracts on demand. \emph{Expandable tool subset} starts with the same Predictor-selected subset, but it allows missing tools to be found through AppWorld's official documentation and added to the subset during execution. \emph{Hierarchical tool awareness} is our proposed core method, guiding tool discovery and recursive decomposition through a capability hierarchy (Table~\ref{tab:ablation-results}).

\begingroup
\setlength{\intextsep}{6pt plus 2pt minus 2pt}
\begin{figure}[!htbp]
\centering
\newsavebox{\ablationtablebox}
\newsavebox{\ablationcaptionbox}
\sbox{\ablationtablebox}{%
\begin{minipage}[t]{0.56\linewidth}
\vspace{0pt}
\centering
\small
\setlength{\abovecaptionskip}{0pt}
\setlength{\belowcaptionskip}{0pt}
\renewcommand{\arraystretch}{1.3}
\setlength{\tabcolsep}{2.5pt}
\begin{tabular}{@{}lccc@{}}
\toprule
Tool awareness & TGC & SGC & ATVS \\
\midrule
Predictor subset & 51.3\% & 32\% & 66.52 \\
Full tool injection & 52.0\% & 22\% & 71.81 \\
Expandable tool subset & 58.7\% & 42\% & 77.78 \\
\textbf{Hierarchical tool awareness} & \textbf{69.3\%} & \textbf{48\%} & \textbf{80.28} \\
\bottomrule
\end{tabular}
\par\nointerlineskip\vspace{5pt}
\captionof{table}{Tool-awareness ablations on the adopted AppWorld Test-C run set. TGC measures task completion, SGC completion of all scenario variants, and ATVS partial verifier progress.}
\label{tab:ablation-results}
\end{minipage}}
\sbox{\ablationcaptionbox}{%
\begin{minipage}[t]{0.415\linewidth}
\vspace{0pt}
\setlength{\abovecaptionskip}{0pt}
\setlength{\belowcaptionskip}{0pt}
\captionof{figure}{Mean tool-context tokens per saved decision (log scale) versus TGC for the four settings in Table~\ref{tab:ablation-results}.}
\label{fig:ablation-context-tgc}
\end{minipage}}
\usebox{\ablationtablebox}\hfill
\begin{minipage}[t][\dimexpr\ht\ablationtablebox+\dp\ablationtablebox\relax][s]{0.415\linewidth}
\vspace{0pt}
\noindent\includegraphics[width=\linewidth,height=\dimexpr\ht\ablationtablebox+\dp\ablationtablebox-\ht\ablationcaptionbox-\dp\ablationcaptionbox-5pt\relax,keepaspectratio]{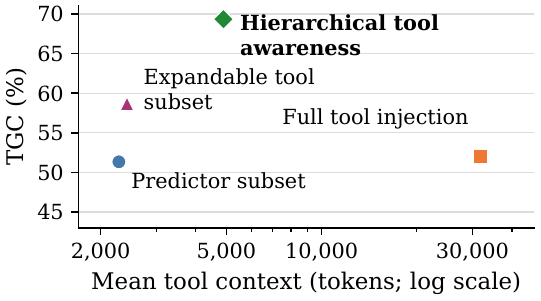}\par
\nointerlineskip\vfill
\noindent\usebox{\ablationcaptionbox}
\par\vspace{0pt}
\end{minipage}
\end{figure}
\endgroup

Full tool injection yields only a slight improvement in task completion over the Predictor subset, suggesting that exposing all tool information does not adequately address the problem: lengthy tool descriptions can distract the agent and complicate tool selection and execution. The expandable tool subset achieves a clearer improvement, supporting on-demand discovery as a way to recover missing tools without placing the entire library in context. Hierarchical tool awareness further improves performance, suggesting that capability hierarchies provide better guidance for task decomposition and execution than flat tool lists. Figure~\ref{fig:ablation-context-tgc} further shows that hierarchical tool awareness substantially improves task execution performance with only a modest increase in tool context.

\noindent\textbf{Tool Exposure and Omission Analysis.}
We further examine whether flat Predictor selection can hide necessary tools. 12\% of the tasks showed strong evidence of a blocking omission, accounting for 24.3\% of the total failed cases. This illustrates a failure mode of committing to a flat subset before execution: a capability that \mbox{was never exposed cannot be selected later.}

For TaReD, we count the distinct tools whose capability information appears in a recorded model context and those for which the detailed executable contract is inspected. Across all runs, capability information was exposed for 7.41\% of the library on average, while detailed contracts were inspected for 2.48\%. The measurements indicate selective exposure in the audited TaReD runs.

%% file: sections/6_discussion_and_conclusion.tex
\section{Conclusion}
TaReD addresses the difficulty of decomposing long-horizon tasks when large tool libraries cannot be exposed in full. It combines on-demand hierarchical tool discovery with capability-guided recursive decomposition. This design aligns subtask boundaries with available capabilities without requiring full-library context. A remaining limitation is the overhead of hierarchical tool discovery: navigating several capability levels can require additional model calls and increase execution time relative to one-shot tool selection. Our current implementation uses a tool cache so that previously queried capability cards and tool information can be reused instead of being discovered repeatedly. This reduces redundant queries but increases the amount of information retained in the working context. The resulting trade-off between discovery overhead and context usage remains open.

%% file: sections/A_appendix.tex
\section{Capability-Tree Construction}

This appendix describes how we transform a conventional flat tool library into the hierarchical capability representation used by TaReD. The construction is performed before task execution; it is therefore a static organization of capabilities rather than a tree inferred separately for each task. Its internal nodes summarize capability categories at progressively finer levels of granularity, while its leaves preserve the original executable tool contracts.

\subsection{From a flat catalogue to a hierarchy}

The input is the public tool catalogue, whose records contain tool identifiers, descriptions, parameter schemas, response information, and any documented preconditions or side effects. We remove the excluded canary field and validate each record's documented identity and required contract fields while preserving its parameters and response schemas. Tools with distinct application semantics or contracts remain separate.

We then assign every tool a single primary path in a capability taxonomy. The path is constructed from progressively more specific functional distinctions: a broad capability domain, an application or business object when needed, an operation scenario, and finally the concrete operation. For example, shopping tools can be organized under a commerce domain, then catalog, cart, or order capabilities, before reaching individual product-search or checkout operations. A new intermediate level is introduced only when it separates meaningful capabilities or execution boundaries; different branches may consequently have different depths. The taxonomy is explicit and versioned, so rebuilding the tree with the same catalogue and taxonomy yields the same primary hierarchy.

The construction creates a root node for the complete library, creates internal capability nodes for the taxonomy paths, and attaches each normalized tool as a leaf under its deepest applicable category. The edges encode parent--child containment and progressively finer functional scope. A tool has one primary parent, which prevents duplicate execution identities and makes descendant counts well-defined. Some capabilities are related across branches, for example authentication may be relevant to several application domains. We represent such relationships with non-tree navigation links rather than copying a tool or treating the link as an execution order. This preserves a tree for hierarchical browsing while allowing the agent to discover a relevant capability outside its current branch \mbox{when the available evidence is insufficient.}

\subsection{Node types and stored information}

The generated catalogue contains three structural node types. The root is a non-executable entry point for the entire library. Category nodes are non-executable internal nodes that summarize a capability region and point to more specific categories or tools. Native-tool nodes are executable leaves that retain one complete tool contract. Composite tools are not created implicitly by the builder: a category is never treated as callable, and a composite operation would require a separately implemented node type and contract.

Every node has a stable identifier, a display name, a node kind, a parent reference when applicable, child references, and a concise semantic summary. Category nodes additionally store the information needed for planning and navigation: the scope or boundary of the region, input and output hints, relevant preconditions, known limitations, representative capability facets, related-node links, source references, and statistics over descendant contracts. These summaries describe what the region can support; they are not a unioned executable schema, and they do not expose the concrete values or current availability of accounts, tokens, or object identifiers.

Each native-tool node stores the exact source identity and executable contract, including the operation path or method, parameter names and types, required and default values, constraints, natural-language description, and structured success or failure response information. Source references and catalogue checksums allow the leaf description to be traced back to the original tool documentation. Runtime state is deliberately excluded from the static tree: values such as access tokens, product identifiers, and current records are obtained from the environment only when the corresponding stage executes.

\subsection{Construction checks and runtime exposure}

After the tree is assembled, we validate that every tool has exactly one reachable primary parent, every category reference resolves, node kinds agree with their executable status, and descendant counts match the attached contracts. We also check that related-node links do not duplicate tools or silently define an execution order, and that each leaf contract agrees with its source record. These checks make catalogue changes fail explicitly \mbox{instead of silently changing a tool's capability region.}

The complete tree is not placed in the initial model context. At runtime, a discovery session begins at the root and exposes category summaries and direct children through bounded browsing or expansion. A category description can provide planning-level evidence without revealing all descendant schemas; a native-tool description reveals the exact contract only when the agent has navigated to that leaf and is ready to bind arguments. Pagination and response budgets bound each observation, while the session records only nodes that have actually been discovered. Thus, the same static tree supports reproducible construction and \mbox{selective, on-demand visibility during recursive planning.}

\subsection{AI-Assisted Curation and Deterministic Construction}
\label{sec:ai-tree-construction}

The capability tree is prepared in two stages. First, an AI agent (Codex) curates a versioned taxonomy from the public AppWorld API documentation. The taxonomy explicitly specifies each category's parent, functional summary, input and output hints, preconditions, limitations, related-category links, and member API identifiers. Categories are organized from broad capability domains toward application or business-object groups and more specific operation types. Each API is assigned to exactly one primary category; cross-cutting relations are recorded as links rather than duplicate assignments. This curation is task-independent: it does not use task instructions, reference solutions, live environment values, or verifier feedback.

Second, a deterministic builder converts the taxonomy and the complete public API-documentation set into the serialized catalogue. It checks that the expected application files are present, verifies each API's documented identity and required contract fields, and removes only the public loader's excluded canary field. It then instantiates non-executable category nodes and executable native-tool leaves, retaining the source API descriptions and exact parameter and response contracts. A traversal from the root assigns depths and derives descendant API counts, application coverage, representative source references, capability facets, and summaries of descendant contract properties. The current catalogue contains 137 category nodes and 457 native-tool leaves.

Finally, validation checks that every documented API is assigned once, each category is nonempty and non-callable, every node is reachable exactly once from the root, parent and depth references agree, and leaf contracts and source pointers match their documented records. The builder also records source and taxonomy hashes in the catalogue version and verifies an integrity hash over the generated artifact. Rebuilding from the same public documentation and taxonomy therefore produces the same tree; changes to either source require a new reviewed taxonomy or a regenerated catalogue. The AI agent curates the static organization, while this deterministic construction and validation process creates the artifact used by the runtime agent.

%% file: sections/B_execution_trace.tex
\section{Illustrative Execution Trace}
\label{sec:illustrative-trace}

This appendix gives a schematic example of how capability discovery, recursive decomposition, and execution fit together. It is illustrative rather than a trace from an evaluated episode; the capability regions and tool identifiers correspond to the catalogue, while the task and returned values are hypothetical.

Suppose the agent receives the goal: ``Read the project deadline from an email and add it as a task in Todoist.'' The goal spans two functional stages: obtaining the deadline from communication data and creating a task in a productivity application. The agent first navigates from the root toward email capabilities. It inspects relevant summaries and children, then refines the goal into (i) locating and reading the relevant message and (ii) creating a Todoist task using the extracted deadline. The second child depends on information produced by the first, so the parent records that dependency instead of asking the second child to infer its inputs.

The first child is handled depth first. Once capability information identifies a suitable operation, the agent retrieves its exact contract, binds the message identifier, and executes it. The returned message content is stored under a stable reference, such as $k_{\mathrm{mail}}$. The parent extracts the deadline and makes it available as a structured value associated with that reference, then closes the first child. The scheduler resumes the second child. Its expected region points toward Todoist task capabilities; if needed, the agent searches that region and selects the concrete \texttt{todoist.create\_task} operation. Only when this stage is ready to execute does the agent inspect the call contract and bind the task title and due date from the task context and stored email result. The returned task identifier and status are saved in the value store, after which the parent checks whether the original goal has been met.

The example highlights two distinctions. Capability summaries guide the choice and granularity of task stages, but they are not executable contracts. The exact contract is bound only after a stage is ready to run. The reference from the email stage preserves the dependency between children without copying the full earlier response into every later planning context. If discovery in the expected region were insufficient, the controller could inspect another relevant capability region before deciding whether to refine or execute the current node.

%% file: main.bbl
\begin{thebibliography}{33}
\providecommand{\natexlab}[1]{#1}
\providecommand{\url}[1]{\texttt{#1}}
\expandafter\ifx\csname urlstyle\endcsname\relax
  \providecommand{\doi}[1]{doi: #1}\else
  \providecommand{\doi}{doi: \begingroup \urlstyle{rm}\Url}\fi

\bibitem[Hao et~al.(2023)Hao, Liu, Wang, and Hu]{hao2023toolkengpt}
Shibo Hao, Tianyang Liu, Zhen Wang, and Zhiting Hu.
\newblock {ToolkenGPT}: Augmenting frozen language models with massive tools
  via tool embeddings.
\newblock In \emph{Advances in Neural Information Processing Systems},
  volume~36, 2023.
\newblock URL
  \url{https://proceedings.neurips.cc/paper_files/paper/2023/hash/8fd1a81c882cd45f64958da6284f4a3f-Abstract-Conference.html}.

\bibitem[Huang et~al.(2022)Huang, Abbeel, Pathak, and
  Mordatch]{huang2022zeroshot}
Wenlong Huang, Pieter Abbeel, Deepak Pathak, and Igor Mordatch.
\newblock Language models as zero-shot planners: Extracting actionable
  knowledge for embodied agents.
\newblock In \emph{Proceedings of the 39th International Conference on Machine
  Learning}, volume 162 of \emph{Proceedings of Machine Learning Research},
  pp.\  9118--9147. PMLR, 2022.
\newblock URL \url{https://proceedings.mlr.press/v162/huang22a.html}.

\bibitem[Huang et~al.(2023)Huang, Xia, Xiao, Chan, Liang, Florence, Zeng,
  Tompson, Mordatch, Chebotar, Sermanet, Brown, Jackson, Luu, Levine, Hausman,
  and Ichter]{huang2023innermonologue}
Wenlong Huang, Fei Xia, Ted Xiao, Harris Chan, Jacky Liang, Pete Florence, Andy
  Zeng, Jonathan Tompson, Igor Mordatch, Yevgen Chebotar, Pierre Sermanet, Noah
  Brown, Tomas Jackson, Linda Luu, Sergey Levine, Karol Hausman, and Brian
  Ichter.
\newblock Inner monologue: Embodied reasoning through planning with language
  models.
\newblock In \emph{Proceedings of the 6th Conference on Robot Learning}, volume
  205 of \emph{Proceedings of Machine Learning Research}, pp.\  1769--1782.
  PMLR, 2023.
\newblock URL \url{https://proceedings.mlr.press/v205/huang23c.html}.

\bibitem[Ichter et~al.(2023)Ichter, Brohan, Chebotar, Finn, Hausman, Herzog,
  Ho, Ibarz, Irpan, Jang, Julian, Kalashnikov, Levine, Lu, Parada, Rao,
  Sermanet, Toshev, Vanhoucke, Xia, Xiao, Xu, Yan, Brown, Ahn, Cortes, Sievers,
  Tan, Xu, Reyes, Rettinghouse, Quiambao, Pastor, Luu, Lee, Kuang, Jesmonth,
  Joshi, Jeffrey, Ruano, Hsu, Gopalakrishnan, David, Zeng, and
  Fu]{ahn2023saycan}
Brian Ichter, Anthony Brohan, Yevgen Chebotar, Chelsea Finn, Karol Hausman,
  Alexander Herzog, Daniel Ho, Julian Ibarz, Alex Irpan, Eric Jang, Ryan
  Julian, Dmitry Kalashnikov, Sergey Levine, Yao Lu, Carolina Parada, Kanishka
  Rao, Pierre Sermanet, Alexander~T. Toshev, Vincent Vanhoucke, Fei Xia, Ted
  Xiao, Peng Xu, Mengyuan Yan, Noah Brown, Michael Ahn, Omar Cortes, Nicolas
  Sievers, Clayton Tan, Sichun Xu, Diego Reyes, Jarek Rettinghouse, Jornell
  Quiambao, Peter Pastor, Linda Luu, Kuang-Huei Lee, Yuheng Kuang, Sally
  Jesmonth, Nikhil~J. Joshi, Kyle Jeffrey, Rosario~Jauregui Ruano, Jasmine Hsu,
  Keerthana Gopalakrishnan, Byron David, Andy Zeng, and Chuyuan~Kelly Fu.
\newblock Do as i can, not as i say: Grounding language in robotic affordances.
\newblock In \emph{Proceedings of the 6th Conference on Robot Learning}, volume
  205 of \emph{Proceedings of Machine Learning Research}, pp.\  287--318. PMLR,
  2023.
\newblock URL \url{https://proceedings.mlr.press/v205/ichter23a.html}.

\bibitem[Karpas et~al.(2022)Karpas, Abend, Belinkov, Lenz, Lieber, Ratner,
  Shoham, Bata, Levine, Leyton-Brown, Muhlgay, Rozen, Schwartz, Shachaf,
  Shalev-Shwartz, Shashua, and Tenenholtz]{karpas2022mrkl}
Ehud Karpas, Omri Abend, Yonatan Belinkov, Barak Lenz, Opher Lieber, Nir
  Ratner, Yoav Shoham, Hofit Bata, Yoav Levine, Kevin Leyton-Brown, Dor
  Muhlgay, Noam Rozen, Erez Schwartz, Gal Shachaf, Shai Shalev-Shwartz, Amnon
  Shashua, and Moshe Tenenholtz.
\newblock Mrkl systems: A modular, neuro-symbolic architecture that combines
  large language models, external knowledge sources and discrete reasoning.
\newblock \emph{arXiv preprint arXiv:2205.00445}, 2022.

\bibitem[Khot et~al.(2022)Khot, Trivedi, Finlayson, Fu, Richardson, Clark, and
  Sabharwal]{khot2022decomposed}
Tushar Khot, Harsh Trivedi, Matthew Finlayson, Yao Fu, Kyle Richardson, Peter
  Clark, and Ashish Sabharwal.
\newblock Decomposed prompting: A modular approach for solving complex tasks.
\newblock \emph{arXiv preprint arXiv:2210.02406}, 2022.

\bibitem[Kim et~al.(2024)Kim, Moon, Tabrizi, Lee, Mahoney, Keutzer, and
  Gholami]{kim2024llmcompiler}
Sehoon Kim, Suhong Moon, Ryan Tabrizi, Nicholas Lee, Michael~W. Mahoney, Kurt
  Keutzer, and Amir Gholami.
\newblock An {LLM} compiler for parallel function calling.
\newblock In \emph{Proceedings of the 41st International Conference on Machine
  Learning}, volume 235 of \emph{Proceedings of Machine Learning Research},
  pp.\  24370--24391. PMLR, 2024.
\newblock URL \url{https://proceedings.mlr.press/v235/kim24y.html}.

\bibitem[Li et~al.(2023)Li, Zhao, Yu, Song, Li, Yu, Li, Huang, and
  Li]{li2023apibank}
Minghao Li, Yingxiu Zhao, Bowen Yu, Feifan Song, Hangyu Li, Haiyang Yu, Zhoujun
  Li, Fei Huang, and Yongbin Li.
\newblock Api-bank: A comprehensive benchmark for tool-augmented llms.
\newblock \emph{arXiv preprint arXiv:2304.08244}, 2023.

\bibitem[Liu et~al.(2024)Liu, Yu, Zhang, Xu, Lei, Lai, Gu, Ding, Men, Yang,
  Zhang, Deng, Zeng, Du, Zhang, Shen, Zhang, Su, Sun, Huang, Dong, and
  Tang]{liu2024agentbench}
Xiao Liu, Hao Yu, Hanchen Zhang, Yifan Xu, Xuanyu Lei, Hanyu Lai, Yu~Gu,
  Hangliang Ding, Kaiwen Men, Kejuan Yang, Shudan Zhang, Xiang Deng, Aohan
  Zeng, Zhengxiao Du, Chenhui Zhang, Sheng Shen, Tianjun Zhang, Yu~Su, Huan
  Sun, Minlie Huang, Yuxiao Dong, and Jie Tang.
\newblock Agentbench: Evaluating llms as agents.
\newblock In \emph{International Conference on Learning Representations}, 2024.

\bibitem[Lu et~al.(2023)Lu, Peng, Cheng, Galley, Chang, Wu, Zhu, and
  Gao]{lu2023chameleon}
Pan Lu, Baolin Peng, Hao Cheng, Michel Galley, Kai-Wei Chang, Ying~Nian Wu,
  Song-Chun Zhu, and Jianfeng Gao.
\newblock {Chameleon}: Plug-and-play compositional reasoning with large
  language models.
\newblock In \emph{Advances in Neural Information Processing Systems},
  volume~36, 2023.
\newblock URL
  \url{https://proceedings.neurips.cc/paper_files/paper/2023/hash/871ed095b734818cfba48db6aeb25a62-Abstract-Conference.html}.

\bibitem[Nakano et~al.(2021)Nakano, Hilton, Balaji, Wu, Ouyang, Kim, Hesse,
  Jain, Kosaraju, Saunders, Jiang, Cobbe, Eloundou, Krueger, Button, Knight,
  Chess, and Schulman]{nakano2021webgpt}
Reiichiro Nakano, Jacob Hilton, Suchir Balaji, Jeff Wu, Long Ouyang, Christina
  Kim, Christopher Hesse, Shantanu Jain, Vineet Kosaraju, William Saunders,
  Xu~Jiang, Karl Cobbe, Tyna Eloundou, Gretchen Krueger, Kevin Button, Matthew
  Knight, Benjamin Chess, and John Schulman.
\newblock {WebGPT}: Browser-assisted question-answering with human feedback.
\newblock \emph{arXiv preprint arXiv:2112.09332}, 2021.

\bibitem[Chen et~al.(2026{\natexlab{a}})Chen, Li, Zhan, and
  Ye]{chen2026schemafill}
Zhi-Kai Chen, Song-Yan Li, De-Chuan Zhan, and Han-Jia Ye.
\newblock {SchemaFill}: Efficient {LLM} tool calling via slot-parallel
  speculative decoding.
\newblock \emph{arXiv preprint arXiv:2610.07086}, 2026{\natexlab{a}}.
\newblock URL \url{https://arxiv.org/abs/2610.07086}.

\bibitem[Paranjape et~al.(2023)Paranjape, Lundberg, Singh, Hajishirzi,
  Zettlemoyer, and Ribeiro]{paranjape2023art}
Bhargavi Paranjape, Scott Lundberg, Sameer Singh, Hannaneh Hajishirzi, Luke
  Zettlemoyer, and Marco~Tulio Ribeiro.
\newblock {ART}: Automatic multi-step reasoning and tool-use for large language
  models.
\newblock \emph{arXiv preprint arXiv:2303.09014}, 2023.

\bibitem[Patil et~al.(2023)Patil, Zhang, Wang, and Gonzalez]{patil2023gorilla}
Shishir~G. Patil, Tianjun Zhang, Xin Wang, and Joseph~E. Gonzalez.
\newblock Gorilla: Large language model connected with massive apis.
\newblock \emph{arXiv preprint arXiv:2305.15334}, 2023.

\bibitem[Prasad et~al.(2024)Prasad, Koller, Hartmann, Clark, Sabharwal, Bansal,
  and Khot]{prasad-etal-2024-adapt}
Archiki Prasad, Alexander Koller, Mareike Hartmann, Peter Clark, Ashish
  Sabharwal, Mohit Bansal, and Tushar Khot.
\newblock {AD}a{PT}: As-needed decomposition and planning with language models.
\newblock In \emph{Findings of the Association for Computational Linguistics:
  NAACL 2024}, pp.\  4226--4252. Association for Computational Linguistics,
  2024.
\newblock \doi{10.18653/v1/2024.findings-naacl.264}.
\newblock URL \url{https://aclanthology.org/2024.findings-naacl.264/}.

\bibitem[Press et~al.(2022)Press, Zhang, et~al.]{press2022selfask}
Ofir Press, Muru Zhang, et~al.
\newblock Measuring and narrowing the compositionality gap in language models.
\newblock \emph{arXiv preprint arXiv:2210.03350}, 2022.

\bibitem[Qin et~al.(2023)Qin, Liang, Ye, Zhu, Yan, Lu, Lin, Cong, Tang, Qian,
  Zhao, Hong, Tian, Xie, Zhou, Gerstein, Li, Liu, and Sun]{qin2023toolllm}
Yujia Qin, Shihao Liang, Yining Ye, Kunlun Zhu, Lan Yan, Yaxi Lu, Yankai Lin,
  Xin Cong, Xiangru Tang, Bill Qian, Sihan Zhao, Lauren Hong, Runchu Tian,
  Ruobing Xie, Jie Zhou, Mark Gerstein, Dahai Li, Zhiyuan Liu, and Maosong Sun.
\newblock Toolllm: Facilitating large language models to master 16000+
  real-world apis.
\newblock \emph{arXiv preprint arXiv:2307.16789}, 2023.

\bibitem[Schick et~al.(2023)Schick, Dwivedi-Yu, Dessì, Raileanu, Lomeli,
  Zettlemoyer, Cancedda, and Scialom]{schick2023toolformer}
Timo Schick, Jane Dwivedi-Yu, Roberto Dessì, Roberta Raileanu, Maria Lomeli,
  Luke Zettlemoyer, Nicola Cancedda, and Thomas Scialom.
\newblock Toolformer: Language models can teach themselves to use tools.
\newblock \emph{arXiv preprint arXiv:2302.04761}, 2023.

\bibitem[Shen et~al.(2023)Shen, Song, Tan, Li, Lu, and
  Zhuang]{shen2023hugginggpt}
Yongliang Shen, Kaitao Song, Xu~Tan, Dongsheng Li, Weiming Lu, and Yueting
  Zhuang.
\newblock {HuggingGPT}: Solving {AI} tasks with {ChatGPT} and its friends in
  {Hugging Face}.
\newblock In \emph{Advances in Neural Information Processing Systems},
  volume~36, 2023.
\newblock URL
  \url{https://proceedings.neurips.cc/paper_files/paper/2023/hash/77c33e6a367922d003ff102ffb92b658-Abstract-Conference.html}.

\bibitem[Chen et~al.(2026{\natexlab{b}})Chen, Zhong, Li, Zhan, and
  Ye]{chen2026peakbench}
Zhi-Kai Chen, Xu-Xiang Zhong, Song-Yan Li, De-Chuan Zhan, and Han-Jia Ye.
\newblock {PeakBench}: Benchmarking resource-aware tool invocation in {LLM}
  agents.
\newblock \emph{arXiv preprint arXiv:2608.24509}, 2026{\natexlab{b}}.
\newblock URL \url{https://arxiv.org/abs/2608.24509}.

\bibitem[Shi et~al.(2025)Shi, Wang, Yan, Ren, Wang, Yin, and
  Ren]{shi2025toolret}
Zhengliang Shi, Yuhan Wang, Lingyong Yan, Pengjie Ren, Shuaiqiang Wang, Dawei
  Yin, and Zhaochun Ren.
\newblock Retrieval models aren't tool-savvy: Benchmarking tool retrieval for
  large language models.
\newblock In \emph{Findings of the Association for Computational Linguistics:
  ACL 2025}, pp.\  24497--24524. Association for Computational Linguistics,
  2025.
\newblock \doi{10.18653/v1/2025.findings-acl.1258}.
\newblock URL \url{https://aclanthology.org/2025.findings-acl.1258/}.

\bibitem[Singh et~al.(2023)Singh, Blukis, Mousavian, Goyal, Xu, Tremblay, Fox,
  Thomason, and Garg]{singh2023progprompt}
Ishika Singh, Valts Blukis, Arsalan Mousavian, Ankit Goyal, Danfei Xu, Jonathan
  Tremblay, Dieter Fox, Jesse Thomason, and Animesh Garg.
\newblock {ProgPrompt}: Generating situated robot task plans using large
  language models.
\newblock In \emph{2023 IEEE International Conference on Robotics and
  Automation (ICRA)}, pp.\  11523--11530, 2023.
\newblock \doi{10.1109/ICRA48891.2023.10161317}.

\bibitem[Song et~al.(2023)Song, Xiong, Zhu, Wu, Qian, Song, Huang, Li, Wang,
  Yao, Tian, and Li]{song2023restgpt}
Yifan Song, Weimin Xiong, Dawei Zhu, Wenhao Wu, Han Qian, Mingbo Song, Hailiang
  Huang, Cheng Li, Ke~Wang, Rong Yao, Ye~Tian, and Sujian Li.
\newblock Restgpt: Connecting large language models with real-world restful
  apis.
\newblock \emph{arXiv preprint arXiv:2306.06624}, 2023.

\bibitem[Tang et~al.(2023)Tang, Deng, Lin, Han, Liang, Cao, and
  Sun]{tang2023toolalpaca}
Qiaoyu Tang, Ziliang Deng, Hongyu Lin, Xianpei Han, Qiao Liang, Boxi Cao, and
  Le~Sun.
\newblock {ToolAlpaca}: Generalized tool learning for language models with 3000
  simulated cases.
\newblock \emph{arXiv preprint arXiv:2306.05301}, 2023.

\bibitem[Trivedi et~al.(2024)Trivedi, Khot, et~al.]{trivedi2024appworld}
Harsh Trivedi, Tushar Khot, et~al.
\newblock Appworld: A controllable world of apps and people for benchmarking
  interactive coding agents.
\newblock \emph{arXiv preprint arXiv:2407.18901}, 2024.

\bibitem[Wang et~al.(2023)Wang, Xu, Lan, Hu, Lan, Lee, and
  Lim]{wang2023planandsolve}
Lei Wang, Wanyu Xu, Yihuai Lan, Zhiqiang Hu, Yunshi Lan, Roy Ka-Wei Lee, and
  Ee-Peng Lim.
\newblock Plan-and-solve prompting: Improving zero-shot chain-of-thought
  reasoning by large language models.
\newblock \emph{arXiv preprint arXiv:2305.04091}, 2023.

\bibitem[Wang et~al.(2025)Wang, Han, Ji, Wang, Baldwin, and
  Li]{wang2025toolgen}
Renxi Wang, Xudong Han, Lei Ji, Shu Wang, Timothy Baldwin, and Haonan Li.
\newblock {ToolGen}: Unified tool retrieval and calling via generation.
\newblock In \emph{International Conference on Learning Representations}, 2025.
\newblock URL
  \url{https://proceedings.iclr.cc/paper_files/paper/2025/hash/b646bdebeb87dfafe2c6f77a63b564e1-Abstract-Conference.html}.

\bibitem[Wei et~al.(2022)Wei, Wang, Schuurmans, Bosma, Ichter, Xia, Chi, Le,
  and Zhou]{wei2022chain}
Jason Wei, Xuezhi Wang, Dale Schuurmans, Maarten Bosma, Brian Ichter, Fei Xia,
  Ed~Chi, Quoc~V. Le, and Denny Zhou.
\newblock Chain-of-thought prompting elicits reasoning in large language
  models.
\newblock \emph{arXiv preprint arXiv:2201.11903}, 2022.

\bibitem[Xu et~al.(2023)Xu, Peng, Lei, Mukherjee, Liu, and Xu]{xu2023rewoo}
Binfeng Xu, Zhiyuan Peng, Bowen Lei, Subhabrata Mukherjee, Yuchen Liu, and
  Dongkuan Xu.
\newblock {ReWOO}: Decoupling reasoning from observations for efficient
  augmented language models.
\newblock \emph{arXiv preprint arXiv:2305.18323}, 2023.

\bibitem[Yao et~al.(2023)Yao, Zhao, Yu, Du, Shafran, Narasimhan, and
  Cao]{yao2023react}
Shunyu Yao, Jeffrey Zhao, Dian Yu, Nan Du, Izhak Shafran, Karthik Narasimhan,
  and Yuan Cao.
\newblock React: Synergizing reasoning and acting in language models.
\newblock \emph{arXiv preprint arXiv:2210.03629}, 2023.

\bibitem[Zheng et~al.(2024)Zheng, Li, Liu, Liu, Luan, and
  Wang]{zheng2024toolrerank}
Yuanhang Zheng, Peng Li, Wei Liu, Yang Liu, Jian Luan, and Bin Wang.
\newblock {ToolRerank}: Adaptive and hierarchy-aware reranking for tool
  retrieval.
\newblock In \emph{Proceedings of the 2024 Joint International Conference on
  Computational Linguistics, Language Resources and Evaluation (LREC-COLING
  2024)}, pp.\  16263--16273. ELRA and ICCL, 2024.
\newblock URL \url{https://aclanthology.org/2024.lrec-main.1413/}.

\bibitem[Zhou et~al.(2023)Zhou, Sch{\"a}rli, Hou, Wei, Scales, Wang,
  Schuurmans, Cui, Bousquet, Le, and Chi]{zhou2023leasttomost}
Denny Zhou, Nathanael Sch{\"a}rli, Le~Hou, Jason Wei, Nathan Scales, Xuezhi
  Wang, Dale Schuurmans, Claire Cui, Olivier Bousquet, Quoc~V. Le, and Ed~H.
  Chi.
\newblock Least-to-most prompting enables complex reasoning in large language
  models.
\newblock In \emph{International Conference on Learning Representations}, 2023.
\newblock URL \url{https://openreview.net/forum?id=WZH7099tgfM}.

\bibitem[Zhou et~al.(2024)Zhou, Xu, Zhu, Zhou, Lo, Sridhar, Cheng, Ou, Bisk,
  Fried, Alon, and Neubig]{zhou2024webarena}
Shuyan Zhou, Frank~F. Xu, Hao Zhu, Xuhui Zhou, Robert Lo, Abishek Sridhar,
  Xianyi Cheng, Tianyue Ou, Yonatan Bisk, Daniel Fried, Uri Alon, and Graham
  Neubig.
\newblock {WebArena}: A realistic web environment for building autonomous
  agents.
\newblock In \emph{International Conference on Learning Representations}, 2024.
\newblock URL \url{https://openreview.net/forum?id=oKn9c6ytLx}.

\end{thebibliography}
